\documentclass{article}

\PassOptionsToPackage{numbers, compress}{natbib}

\usepackage[final]{neurips_2019_ml4ps}

\usepackage[utf8]{inputenc} 
\usepackage[T1]{fontenc}    
\usepackage{hyperref}       
\usepackage{url}            
\usepackage{booktabs}       
\usepackage{amsfonts}       
\usepackage{nicefrac}       
\usepackage{microtype}      

\usepackage{float}
\usepackage{graphicx}
\usepackage{multirow}
\usepackage{algorithm}
\usepackage{amsmath}
\usepackage{algpseudocode}

\newcommand{\RR}{\mathbb{R}}

\title{Efficient Parameter Sampling for Neural Network Construction}

\author{%
    Drimik Roy Chowdhury\thanks{This work was done in the University of Oxford.} \\
    Department of Mathematics\\
    University of Michigan\\
    \texttt{drimikr@umich.edu} \\
    \And
    Muhammad Firmansyah Kasim \\
    Department of Physics \\
    University of Oxford \\
    \texttt{muhammad.kasim@physics.ox.ac.uk} \\
}

\begin{document}

\maketitle

\begin{abstract}
    The customizable nature of deep learning models have allowed them to be successful predictors in various disciplines. These models are often trained with respect to thousands or millions of instances for complicated problems, but the gathering of such an immense collection may be infeasible and expensive. However, what often occurs is the pollution of redundant information from these instances to the deep learning models. This paper outlines an algorithm that dynamically selects and appends instances to a training dataset from uncertain regions of the parameter space based on differences in predictions from multiple convolutional neural networks (CNNs). These CNNs are also simultaneously trained on this growing dataset to construct more accurate and knowledgable models. The methodology presented has reduced training dataset sizes by almost 90\% and maintained predictive power in two diagnostics of high energy density physics.
\end{abstract}

\section{Introduction}
Time, material resources and human labor required to administer experiments are often not scalable for a deep understanding of the possible results based on varying input combinations -- hence, the need for models such as neural networks to learn these relationships. An ideal model would serve to predict the outcomes of physical experiments for all if not most possible conditions. However, it may be inconceivable to collect thousands or millions of training instances for the model as performing these many experiments may in fact leave the idea of building a model infeasible and thus, pointless for actual physical application. What occurs often is that training datasets are riddled with redundancies that do not provide any new information to these models and in fact impair generalization strength \cite{training_redundancy_1,training_redundancy_2}. If there existed a procedure to select training instances based on current knowledge of the parameter space -- thereby dynamically building a training dataset -- this would possibly lessen costs of experimentation by choosing well-crafted inputs that provide unknown information of the results and train an accurate model from a small yet sufficiently informative training dataset. 

This paper introduces an effective methodology to achieve this goal and is tested in the context of two diagnostics in plasma physics (background provided in Section \ref{sec:diagnostics_background}) -- inelastic X-ray Thomson scattering (XRTS) \cite{m6_lee-2009-xrts,m7_kritcher-2008-xrts,m8_valenzuela2018measurement-xrts} and X-ray emission spectroscopy (XES) \cite{m10_regan2013hot-xes,m11_ciricosta2017simultaneous-xes} -- by training multiple convolutional neural networks simultaneously and using their impressions on the outputs as a form of uncertainty to carefully select instances to train on next. This approach constructs a feedback loop on training models and selecting training instances based on what the models are unsure of. We note that the algorithm presented has produced effective training datasets for deep learning models that are approximately 90\% smaller in size than those of current models -- with XRTS a 3 dimensional problem and XES a 10 dimensional.

\section{Algorithm}
\label{sec:algo}
The overarching goal of the algorithm is to produce a suitable training dataset from a continuous parameter space by selecting new points to add based on the information provided by instances already collected. Orthogonal sampling is often a statistical method used in generating a pseudo-random sample of a multidimensional space, popular in numerical integration for Monte-carlo applications and for computer experiments \cite{tang1993_orthogonalsampling}. The basic premise is that the parameter space is divided into \emph{blocks} and points are generated within each block to create a representative sample across the input space. We use this idea to construct a systematic scheme to represent different regions of space to which we associate uncertainty with our predictions. Suppose our parameter space is the $n$-unit cube of $\RR^n$ (inputs can be scaled if necessary). We divide each of the standard intervals of the axes of our input space into finer, disjoint intervals and together, a block representation is formed. The algorithm should store instances to the blocks they are contained in (i.e. the dataset) with their corresponding uncertainties. The uncertainty figures aggregated in each block provide a general idea on how confident the models are on the instances within the block, which the algorithm can use to navigate the input space; in our study, we calculate the average uncertainty as a heuristic for this aggregation with a decaying factor attributed to the uncertainties of older instances collected.

In short, the knowledge of uncertainty is gathered by multiple deep learning models that are concurrently trained by the algorithm with the key idea being that points where the models disagree on the most are points of most uncertainty in predictions. We note that the inspiration for this approach was from Section 2.2.4 of \cite{algo_inspiration} to select instances based on highest variance across ensemble members of reward predictors in Deep Reinforcement Learning. With $l$ models employed, the standard variances along the coordinates of the $l$ outputs of an input $\mathbf{x} \text{ in the } n\text{-unit cube} \text{ of } \RR^n$ is computed. The uncertainty attributed to $\mathbf{x}$ across all models used in this paper is simply the sum of these aforementioned coordinate variances. This allows us to select instances to add to the dataset at each epoch of the algorithm from some set of points. 

Now it remains to define some criteria to select blocks for an epoch of the algorithm as it may be infeasible to always consider all of them (e.g. a simple 2 block division across each parameter of XES results in 1024 blocks). The whole basis of the algorithm is to choose blocks that the models identify as uncertain regions. However, the issue by always following this protocol is that other blocks that haven't had adequate attention could reflect regions of greater uncertainty than where the models believe they are most uncertain in. In this paper, we select random blocks or blocks with the lowest number of points for XRTS and XES, respectively, at the start of the algorithm and every so often as a solution. In addition, with uncertainty solely based on the perspective of the models of the algorithm, there may be instances that the models are ostensibly confident on, but it may be the case that the true output is very different from what all the models believe. To combat this possible bias, we use a decaying $\epsilon$-schedule of probabilities across the epochs of the algorithm to either select random or most uncertain instances from a set of points in the parameter space to add to the dataset. The decaying aspect arises because the models are ideally less biased with regards to this criteria as the algorithm progresses and are appropriately choosing points of highest uncertainty. 

Towards the middle stages of the algorithm, training is performed in every epoch with respect to a fixed number of the most recently added instances to the dataset and periodically on the entire dataset. The reason for this is to introduce robustness to the models' uncertainty calculations because as we include more training instances, the hope would be that the most recently added points are a more accurate representation of regions of uncertainty in space and therefore, reveal most informatively where the models have a lack of confidence on inputs. The pseudo-code is displayed in Algorithm \ref{algo:pseudo_code}.

\begin{algorithm}
\caption{Sample Selection and Neural Network Construction}
\label{algo:pseudo_code}
\begin{algorithmic}[1]

\State Create block division 
\State Gather few instances by Latin hybercube sampling and store them with no uncertainty
\State Train all models

\For{each epoch}
    \State Select blocks by highest average uncertainty or exploration protocol
    \State $\mathbf{X} \gets$ Generate points in each block
    \State Calculate uncertainty on each instance of $\mathbf{X}$
    \State Select most uncertain or random instances from $\mathbf{X}$ based on probability $\epsilon$
    \State Store instances (from previous step) and corresponding uncertainties to their blocks
    \State Train all models
\EndFor
\end{algorithmic}
\end{algorithm}

\section{Deep learning model architecture}
\label{sec:cnn_architecture}

\begin{figure}
    \centering
    \includegraphics[width=\linewidth]{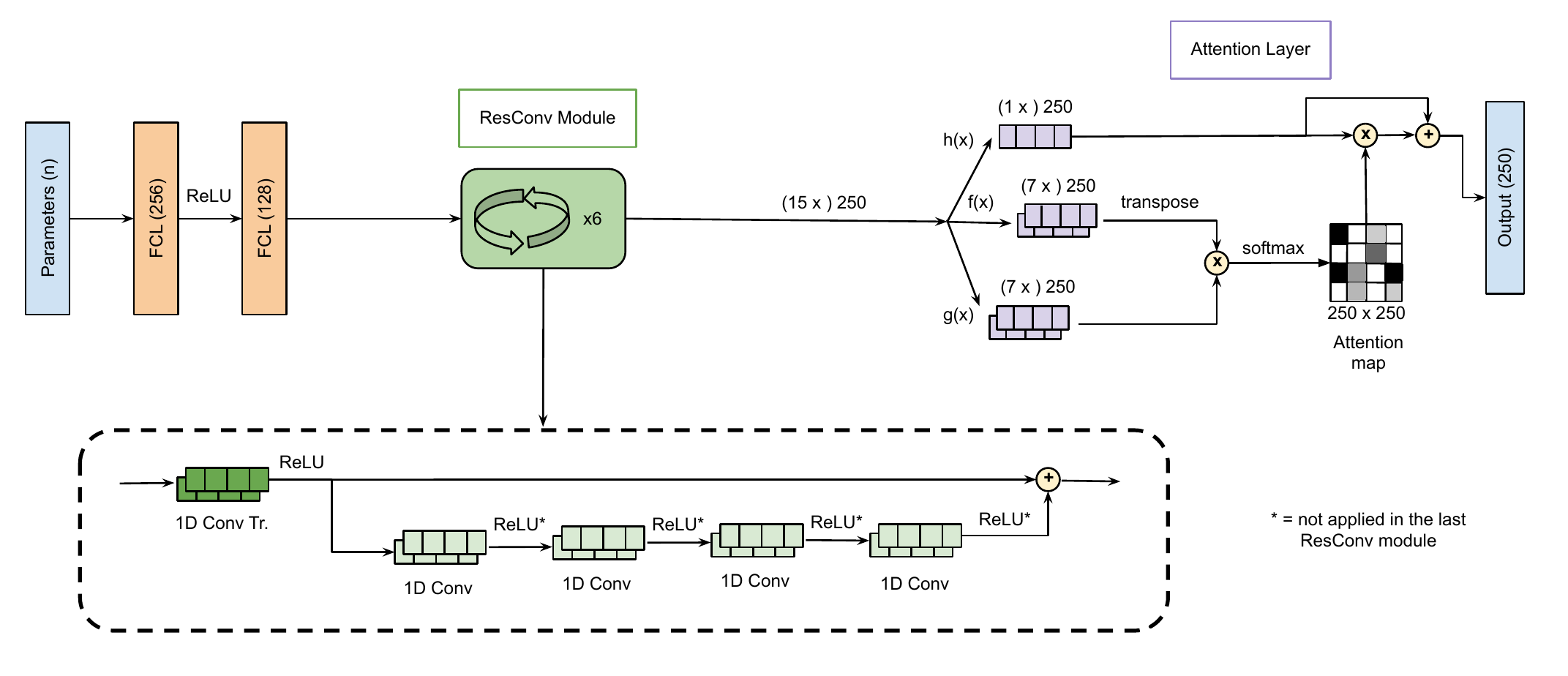}
    \caption{The deep learning model architecture consists of fully connected layers, transposed convolutional layers, convolutional layers, and an attention layer. The $\bigoplus$ and $\bigotimes$ denote skip connections and matrix multiplication, respectively. Note that ($n$ x) $m$ means there are $n$ channels, each $m$ entries long and $n$ x $m$ denotes the standard $n$-by-$m$ matrix. The ResConv module repeats 6 times beginning with an input of 32 channels of length 4.}
    \label{fig:cnn_architecture}
\end{figure}

Figure \ref{fig:cnn_architecture} outlines the model structure implemented in this paper. The convolutional layers are all created with a kernel size of 3, stride of 1, and zero padding of 1. Similarly, the transposed convolutional layers are associated with a kernel size of 2, stride of 1 and no zero padding for the first 4 ResConv modules (defined in Figure \ref{fig:cnn_architecture}) followed by zero padding of 1. The attention layer is inspired from \cite{m19_sagan} in order to capture global dependencies of the output signal \cite{m30_vaswani2017attention,m31_wang2018-non-local-neural-network}. The function $h$ in Figure \ref{fig:cnn_architecture} takes the first of the 15 input channels and $f$ and $g$ splits the remaining channels evenly. We see that $h$ acts as the convolutional information aspect of the attention layer whereas $f$ and $g$ create the attention map that acts on $h$. The attention layer has two learnable parameters: one for the skip connection from $h$, $\beta$, and the other from the attention map applied to $h$, $\alpha$. The neural network training was performed with the ADAM optimizer \cite{m37_kingma2014adam} to minimize the mean squared error.

\section{Results}
The block division for both XRTS and XES was of 2 partitions across each parameter, resulting in 8 and 1024 blocks, respectively. Three models were used in both tests. To assess the degree of success of the algorithm, validation and test datasets of 3000 and 4000 instances, respectively, were generated for each test case with random parameter values over a predefined range that are valid in each context. The algorithm ran for about 9 and 24 hours for XRTS and XES, respectively, on a Titan X GPU card.

\begin{figure}
\begin{minipage}{.5\textwidth}
    \includegraphics[width=\linewidth]{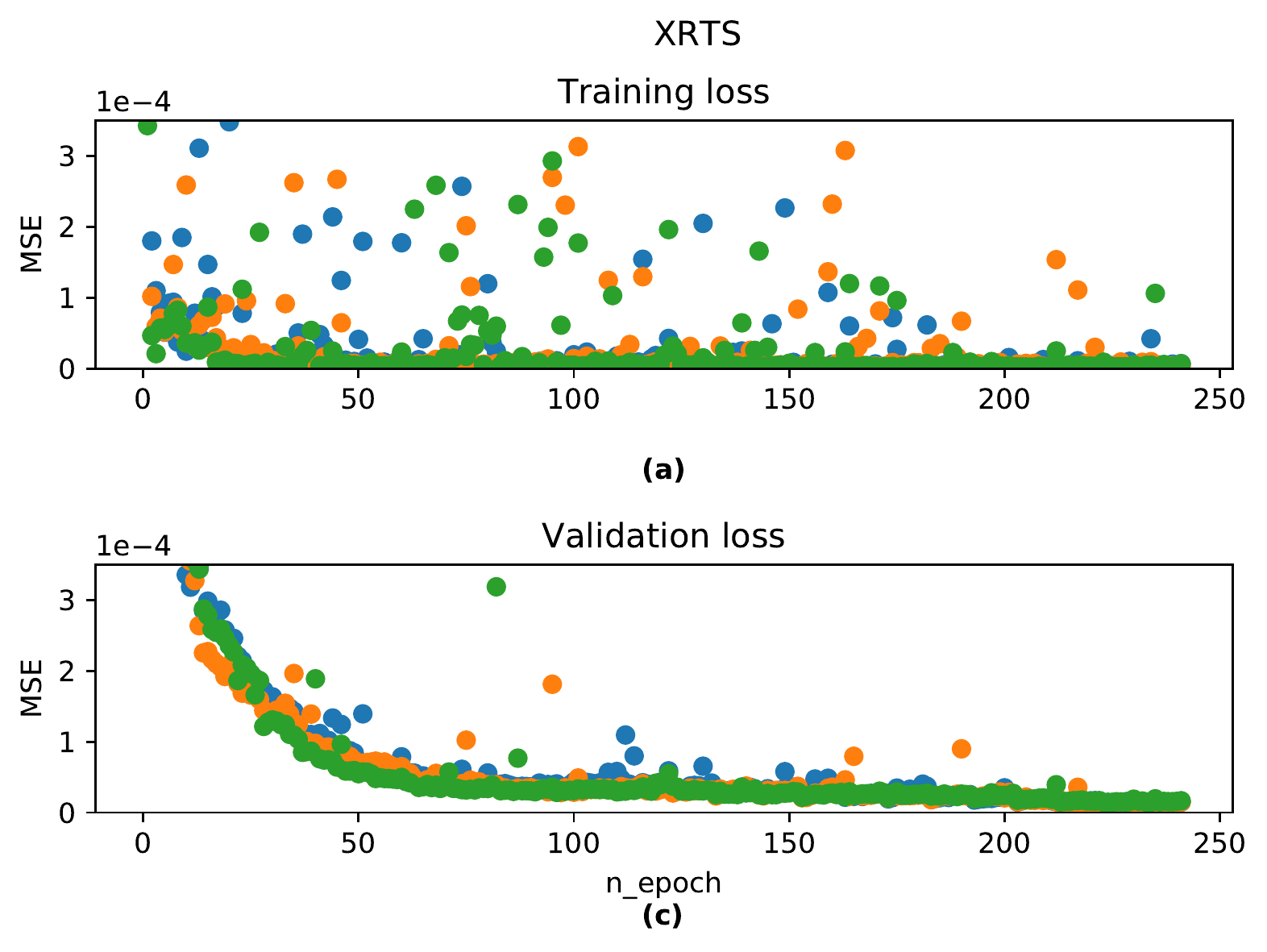}
\end{minipage}%
\begin{minipage}{.5\textwidth}
    \includegraphics[width=\linewidth]{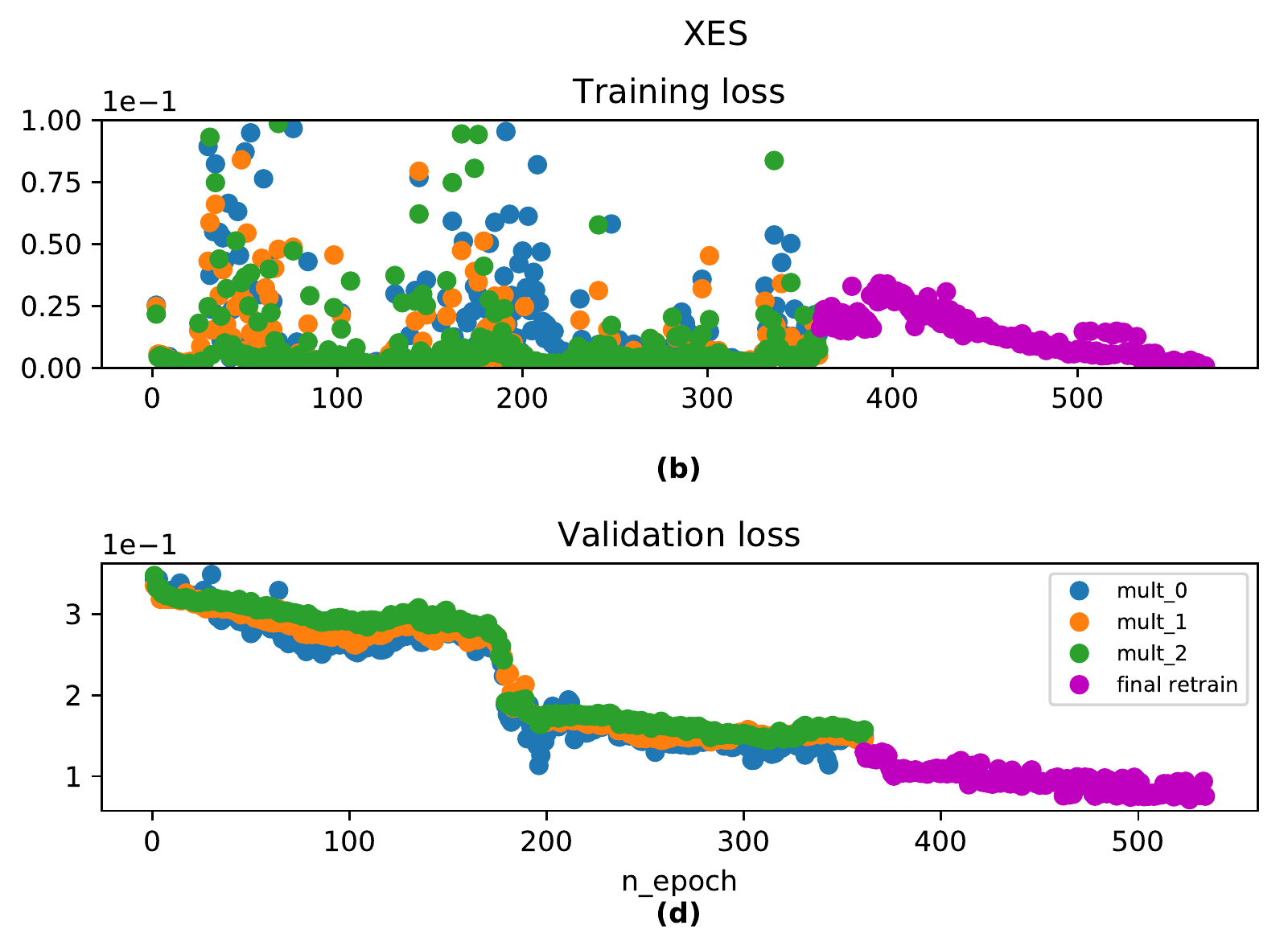}
\end{minipage}
\caption{XRTS and XES MSE losses during each epoch. The three models used for each test case in the algorithm are the three labeled models. The final retraining in XES is performed with the model weights of the lowest validation loss over the dataset created by the algorithm.}
\label{fig:xrts_xes_losses}
\end{figure}
\begin{figure}
    \centering
    \includegraphics[width=\linewidth]{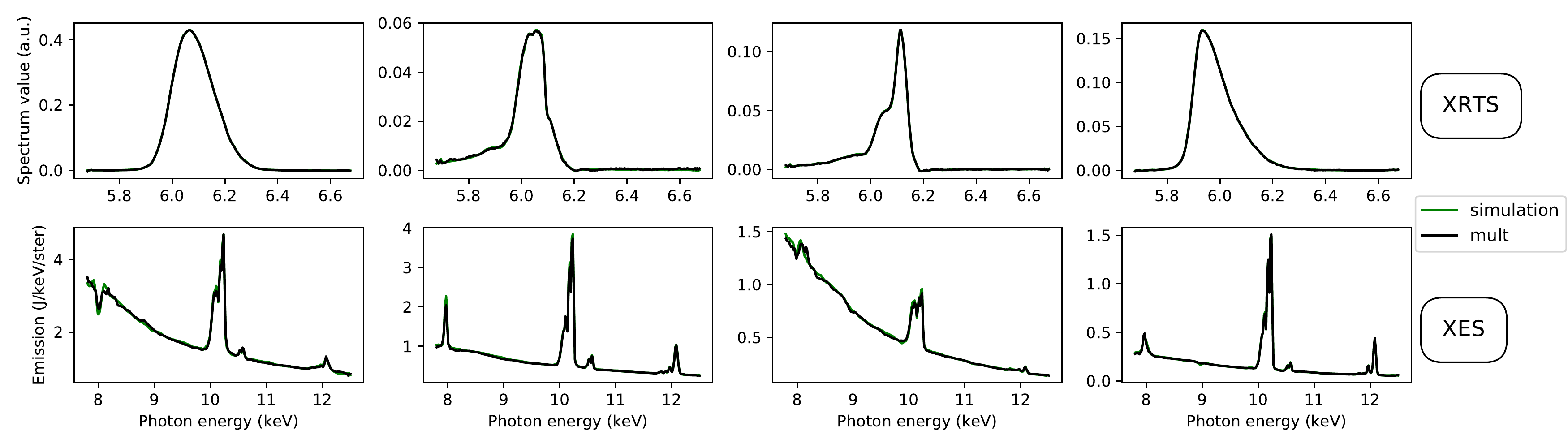}
    \caption{Predictions from XRTS and XES models compared to true outputs. We can see that the model predictions are well in agreement with the outputs.}
    \label{fig:xrts_xes_preds}
\end{figure}

The final validation and test MSE losses for the XRTS case corresponding to the model of the lowest validation loss were 1.1836 x $10^{-5}$and 1.9273 x $10^{-5}$, respectively. For XES, the corresponding losses were 7.273 x $10^{-2}$ and 7.674 x $10^{-2}$, respectively. First and foremost, we note that in Figure \ref{fig:xrts_xes_losses}(a) and (b), the models are seemingly overfitting their respective training datasets. An understandable factor for this occurrence could be the extensive training on these models with respect to the small training datasets. A critical indicator of the success of the uncertainty usage is that for both the XRTS and XES test case, the models repeatedly get ``surprised" when we analyze Figures \ref{fig:xrts_xes_losses}(a) and (b). The frequent increases in training losses of the models signify that the newly added instances provide possibly unfamiliar information to these models of the parameter space.

Alternatively, the general trends of validation losses in Figure \ref{fig:xrts_xes_losses}(c) and (d) are that they decrease as the algorithm progresses but the rates of decay are quite different. By epoch 50, XRTS models explore a large portion of the uncertain regions in its parameter space reaching an MSE validation loss of approximately 6.25 x $10^{-5}$ across all models; the remaining 190 epochs slowly but generally reduce the validation loss until a final value of 1.3 x $10^{-5}$. A reason for this trajectory of validation losses might be because of the relatively small number of compact blocks to investigate, which allows the models to obtain an intuitive idea of the variation of the outputs associated with the entire parameter space early in the algorithm. In contrast, the many more blocks in XES causes the algorithm to possibly require more time to find which part of space (i.e. which blocks) the models are most uncertain in. This is most visible with the drastic decrease of the validation losses by 37.9\% across all 3 models to 1.678 x $10^{-1}$ over the course of epochs 160 to 200. Overall, it may have been the case that if the algorithm was allowed more time to survey the parameter space, another rich source of information may have been located. In addition, the algorithm may also be punishing the XES models for their incorrect perceptions of uncertainties with the slight increase of validation loss before the aforementioned drop. In XES, the final 200 epochs train the model weights of the lowest validation loss of 1.208 x $10^{-1}$ on the entire the training dataset generated in the first 360 epochs, decreasing the MSE validation by approximately 39.8\%. In summary, the XES models spend significantly more time exploring than the XRTS models before they finally recognize where to focus their attention with regards to uncertain inputs. The final XRTS training dataset contained 774 points, which is roughly 11.06\% of the current model's. Similarly, the XES dataset included 912 instances or 13.03\% of its current model's.

In conclusion, the algorithm provided a methodical collection of a training dataset based on uncertainty in predictions and trains deep learning models to learn from this information. This has reduced the amount of training data by an order of magnitude without severely impacting performance.

\section{Diagnostics Background}\label{sec:diagnostics_background}
XRTS and XES were selected as case studies because these methodologies are applied frequently in plasma physics experiments \cite{m6_lee-2009-xrts,m7_kritcher-2008-xrts,m8_valenzuela2018measurement-xrts, m36_inst_xes, m37_inst_xes2}. 

\textbf{X-ray Thomson scattering (XRTS)}
XRTS is a procedure to probe the temperature and density of plasma by emitting an X-ray pulse into the plasma and measuring the spectrum of the scattered pulse with respect to an angle. This spectrum can be identified once the angle and the condition of the plasma is known \cite{m39_xrts_angle}. This paper deals with the case of only 3 parameters -- temperature, ionization, and density -- are retrieved from the scattering of an X-ray at 90 degrees on a Beryllium sample. A single simulation takes approximately 10-20 seconds for completion.

\textbf{X-ray emission spectroscopy (XES)}
XES functions as another method to probe plasma conditions by measuring the emitted spectrum of X-ray and comparing results with simulations or theoretical models. We consider the case of an implosion condition \cite{m10_regan2013hot-xes} from a Germanium emission spectrum with the model outlined in \cite{m11_ciricosta2017simultaneous-xes}. A simulation takes approximately 10-20 minutes to execute.

\small
\bibliographystyle{unsrt}
\bibliography{ref}



\end{document}